\documentclass[10pt,twocolumn,letterpaper]{article}

\usepackage{iccv}
\usepackage{times}
\usepackage{epsfig}
\usepackage{graphicx}
\usepackage{amsmath}
\usepackage{amssymb}
\usepackage{booktabs} 
\usepackage{comment}
\usepackage{multirow}

\usepackage[pagebackref=true,breaklinks=true,letterpaper=true,colorlinks,bookmarks=false]{hyperref}

\iccvfinalcopy 

\begin{document}

\title{Online Overexposed Pixels Hallucination in Videos with Adaptive
Reference Frame Selection}


\author{Yazhou Xing$^{1}$ \quad Amrita Mazumdar$^2$ \quad Anjul Patney$^2$ \quad Chao Liu$^2$ \quad Hongxu Yin$^2$ \\
Qifeng Chen$^1$ \quad Jan Kautz$^2$ \quad Iuri Frosio$^2$
\and
$^1$HKUST \quad $^2$NVIDIA
}

\maketitle

\begin{abstract}
Low dynamic range (LDR) cameras cannot deal with wide dynamic range inputs, frequently leading to local overexposure issues. 
We present a learning-based system to reduce these artifacts without resorting to complex acquisition mechanisms like alternating exposures or costly processing that are typical of high dynamic range (HDR) imaging.
We propose a transformer-based deep neural network (DNN) to infer the missing HDR details.
In an ablation study, we show the importance of using a multiscale DNN and train it with the proper cost function to achieve state-of-the-art quality.
To aid the reconstruction of the overexposed areas, our DNN takes a reference frame from the past as an additional input.
This leverages the commonly occurring temporal instabilities of autoexposure to our advantage: since well-exposed details in the current frame may be overexposed in the future, we use reinforcement learning to train a reference frame selection DNN that decides whether to adopt the current frame as a future reference.
Without resorting to alternating exposures, we obtain therefore a causal, HDR hallucination algorithm with potential application in common video acquisition settings. Our demo video can be found~\href{https://drive.google.com/file/d/1-r12BKImLOYCLUoPzdebnMyNjJ4Rk360/view?usp=sharing}{here}. 
\end{abstract}

\section{Introduction}
\label{sec:introduction}
Consumer-grade video acquisition devices often achieve a trade-off between cost and quality and therefore do not support HDR capture. 
Although autoexposure~\cite{Lia07, Su15, Bernacki2020AutomaticEA, Onzon_2021_CVPR} can drive the exposure time and sensor gain to control the overall frame brightness and capture the most relevant scene details in LDR, its time response is not immediate, leading to globally over- or under-exposed frames during light transitions. As a consequence, amateurs who lack professional light control knowledge, may capture overexposed frame sequences, especially when encountering a sudden illuminance switch such as moving from inside to outside. In this paper, we focus on hallucinating the overexposed details for consumer-grade cameras, in an online manner, and without any additional camera control.

\begin{figure}
\center
\includegraphics[width=0.47\textwidth]{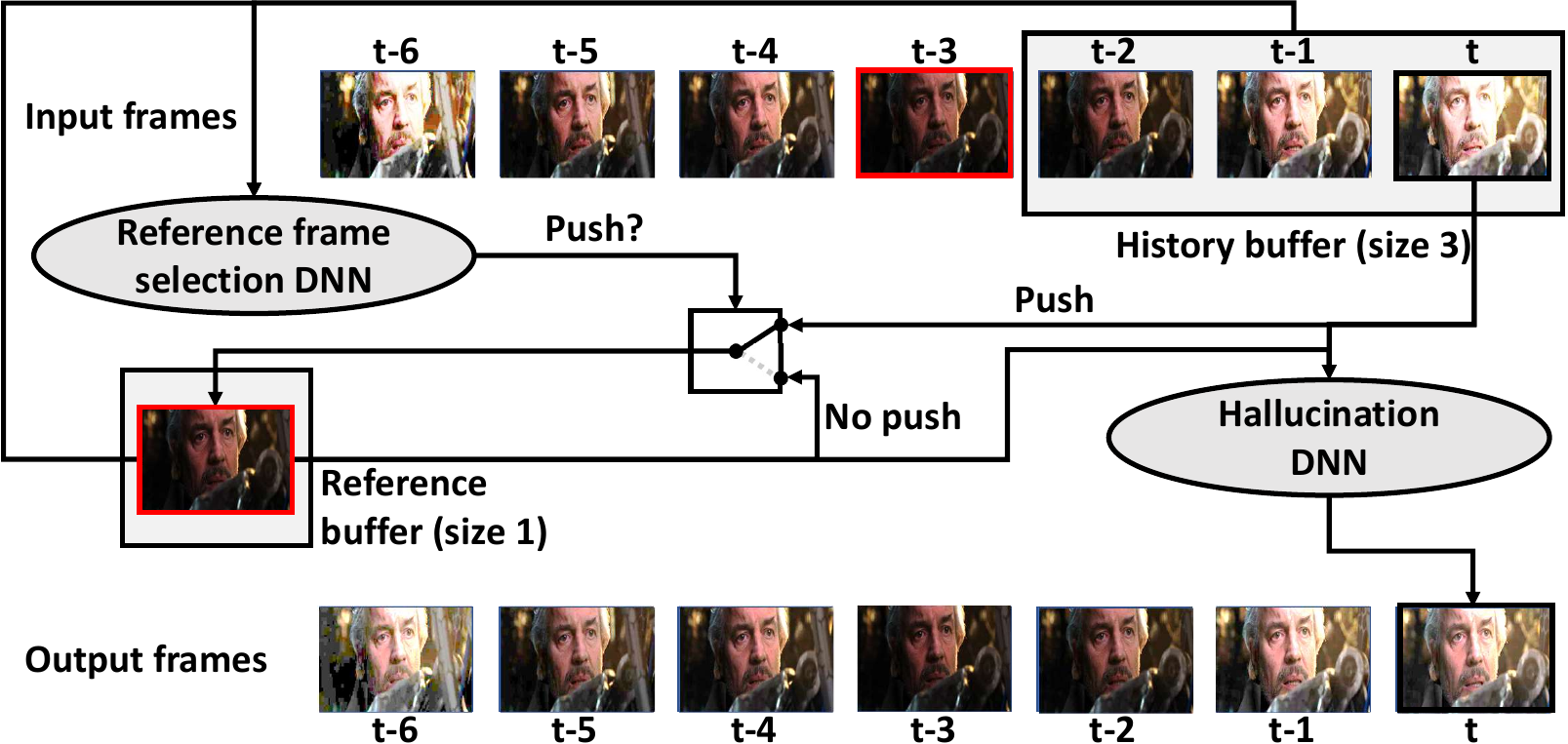}
\caption{We hallucinate saturated details in overexposed areas in videos using a reference frame selection DNN, trained with RL, that leverages the autoexposure oscillations. It processes the three most recent frames (history buffer) and the reference frame (in red) and decides whether to push the current frame into the reference buffer. Our transformer-based hallucination DNN takes the reference and current frame in input to recreate the overexposed details.}
\label{fig:teaser}
\end{figure}
\begin{figure*}
\center
\begin{tabular}{c@{\hspace{0.1mm}}c@{\hspace{0.1mm}}c@{\hspace{0.7mm}}}
\rotatebox{90}{\small \hspace{1mm} Output \hspace{4mm} Input \hspace{3.5mm} Reference \hspace{2mm} Mask}&
\includegraphics[width=0.97\textwidth]{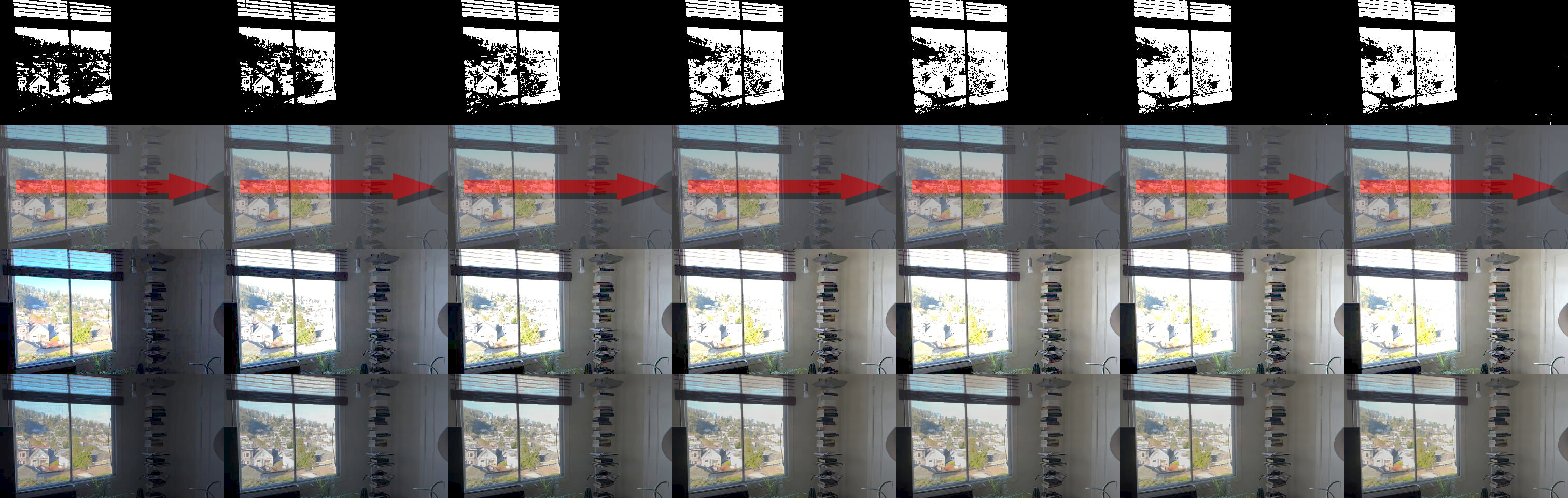}
\end{tabular}
\caption{Results of our overexposure hallucination system on real-world amateur-captured videos. From top to bottom, we show the overexposure mask, the selected reference frame, the current frame, and our reconstruction, respectively. The arrows indicate that the reference frame is kept unaltered for processing the next frame; the frame at the very beginning is selected as the reference frame. Our system automatically picks the most suitable reference frame on-the-fly; this shows a good trade-off between overexposure level and content similarity with the current frame.}
\label{fig:teaser}
\end{figure*}
The filling of overexposed areas also lies at the heart of the HDR reconstruction problem, where HDR details are inherited from one or more LDR (\emph{reference}) frames. It also fits under the umbrella of inpainting or hallucination, where pixels are reconstructed to guarantee semantic and visual plausibility for human observers. 
Here we combine these two philosophies into a unique system.

Because of the quality of the viewing experience provided by HDR images and videos, there have been many approaches to reconstructing overexposed information from LDR images~\cite{Eil17_hdrcnn, endo2017deep, Liu20_singlereverse, San20_singlehdr, MertensKR09, COGALAN202257, Kal19, chen2021hdr}. Overexposed details hallucinated from a single LDR source~\cite{Eil17_hdrcnn, endo2017deep, Liu20_singlereverse, San20_singlehdr} may look plausible, but there is no promise to realistic scenes since they are only \emph{imagined} by the \emph{prior} learned by the DNN. The fake details may even be magnified in large saturated areas~\cite{Eil17_hdrcnn, Liu20_singlereverse}. Another approach~\cite{MertensKR09, COGALAN202257, Kal19, chen2021hdr} uses multiple exposures and no imagination capability: the reconstructed details are generally high fidelity as they are essentially copied and pasted from similar frames with different exposures. However, this requires full camera control (\eg{}, alternating exposures~\cite{COGALAN202257, Kal19}) and it is often computationally intensive because of frame realignment~\cite{Kal19}. 
The hallucination of saturated pixels in videos has also connections with video inpainting~\cite{Elh19, Iiz17_localglobal, Liu18_partial, Jia18_contextual}. The first solutions were aimed at consistent propagation of texture and structures~\cite{Elh19}. Semantic consistency was later enforced~\cite{Elh19, Iiz17_localglobal, Liu18_partial, Jia18_contextual}:
state-of-the-art methods include the attention mechanism provided by transformers~\cite{Li22_endtoend, Liu21_FuseFormer, Yan20_sttn, Ren22, liu2022ghost}. However, when dealing with overexposure, the hallucination network also has to deal with the problem of merging data from frames with different exposures.

Here, we propose a novel and effective solution for \emph{online} overexposed pixel hallucinations in videos, without any form of camera control. We leverage (for the first time, to the best of our knowledge) the uncontrolled temporal exposure variations, that offer the opportunity to select one well-exposed frame as a reference within the video itself, to hallucinate saturated details in overexposed frames. Our solution consists of two parts: the hallucination model and the reference frame selection. We carefully design the reference-based hallucination model with attention to identifying semantic connections not only within the current frame but also towards a past frame used as a reference. For the frame selection problem, we show the advantage of selecting a single, good reference frame and tackle the problem with a DNN trained through reinforcement learning (RL). Our selection strategy uses only frames from the past, enabling our method to run online with minimum latency. Overall, our technique resembles the traditional reference-based HDR setting, while simultaneously leveraging the hallucination capabilities of transformers. Our contributions include:

\begin{itemize}
    \item firstly addressing the problem of online overexposed detail hallucination in videos for HDR imaging with deep learning, where no camera control apart from autoexposure is required;
    \item introducing the first reference-based transformer DNN to hallucinate overexposed details in LDR videos that achieves state-of-the-art results, and performing an ablation study of its features (the multiscale architecture, the position embedding, and the training loss); 
    \item demonstrating the possibility of using a single, good reference frame rather than using a large set of frames in input for reference-based hallucination;
    \item introducing a reference frame selection DNN and showing that it is possible to leverage the exposure oscillations in video sequences and train the reference frame selection DNN using RL to learn an optimal, online reference frame selection strategy.
\end{itemize}

\section{Related Work}
\label{sec:relatedwork}

Hallucinating image details requires \emph{self-consistency}: the new pixels must match the image content both visually and semantically, and look \emph{plausible} at visual inspection.
In videos, \emph{cross-consistency} among frames has to be enforced to guarantee temporal stability.
In the traditional HDR setting, the use of a stack of differently exposed frames allows the reconstruction to be not simply plausible, but \emph{as adherent as possible to the reality}.
Here we list the characteristics, pros and cons of recent hallucination and HDR algorithms,  taking into account the concepts of self-consistency, cross-consistency, and adherence to reality.

\textbf{HDR algorithms.} Some HDR algorithms recover HDR details in LDR images~\cite{banterle2006inverse, wang2007high, Eil17_hdrcnn, endo2017deep, eilertsen2019single, Liu20_singlereverse, San20_singlehdr}; although tone mapping of HDR data into an LDR domain could be taken in account ~\cite{meylan2007tone, Ou22_tonemapping}, like others~\cite{Eil17_hdrcnn, Liu20_singlereverse} we prefer working in the linear domain.
High quality HDR methods exploit complex camera control patterns (\eg{}, alternating exposures), but require camera control and costly frame alignment and merging~\cite{masia2009evaluation, Tur15, mccann2011art, Has16, Kal19,chen2021hdr,liu2021adnet}.
Since they use frames from the neighboring window, stack-based HDR methods can achieve semantic adherence to reality. 
Turning an LDR frame into HDR~\cite{Mar18_expandnet} is computationally light and does not require camera control, but requires hallucinating the missing details.
Rempel et al.~\cite{rempel2007ldr2hdr} design a system for on-the-fly reverse tone mapping for videos. However, they cannot recover the missing saturated detail but only aim at enlarging the contrast and increasing the luminance of saturated regions.
Lastly, the lack of large HDR datasets is problematic for HDR algorithms, obliging to pretrain DNNs on LDR inpainting~\cite{San20_singlehdr} or to simulate HDR data~\cite{Eil17_hdrcnn}.

\textbf{Single frame vs. video hallucination.} We distinguish between algorithms recreating missing details in a single frame~\cite{Eil17_hdrcnn, Mar18_expandnet, San20_singlehdr, Liu20_singlereverse, xing2021invertible, Liu18_partial, Jia18_contextual, Iiz17_localglobal, Li22_mat} from those targeting videos~\cite{Yan20_sttn, Liu21_FuseFormer, Li22_endtoend} and thus requiring temporal stability: \eg{}, one of the issues of~\cite{Liu20_singlereverse} is flickering.
Furthermore, many video algorithms get frames from both the past and the future and are therefore unsuitable for online, real-time applications where high latency is not admissible.

\textbf{Reference frame vs. pure hallucination.} By using self-attention (dictating that hallucinated pixels must be semantically consistent with the frame~\cite{Li22_mat}), single frame hallucination methods~\cite{Eil17_hdrcnn, Liu20_singlereverse, San20_singlehdr} recreate plausible content, but not necessarily adherent to reality.
The adoption of a reference frame can overcome this issue, as already demonstrated for instance in super-resolution~\cite{Yan20_learningsuperres}.
Methods forcing consistency within entire video sequences~\cite{Yan20_sttn, Liu21_FuseFormer, Li22_endtoend, lei2020blind}, as well as stack-based HDR methods, remain however computationally intensive.
Ours is the only method to explicitly use cross-attention to connect the hallucinated content with one reference frame: it aims at restoring the real signal, not simply a plausible alternative. 
A copy-and-paste mechanism for video interpolation has already been proposed~\cite{Lee19}: it identifies the relevant reference patches, realigns and copies them in the current frame. However, as opposed to our transformer-based DNN, it does not handle HDR data or large perspective changes between the reference and current frame.

\textbf{Reference frame selection. } The choice of working with a limited number of reference frames is computationally cheaper, but it also requires solving an additional problem: how to identify them.
In some cases, this is done trivially (\eg{}, using constant intervals in video compression~\cite{Tri21, Hu2021FVCAN}).
Researchers investigated automatic identification of the most informative frames for action and video recognition~\cite{Wu_2019_CVPR, meng2021adafuse, Gowda_Rohrbach_Sevilla-Lara_2021}. We develop frame selection based on RL and design an online system that uses only past frames, thus achieving low latency for the video streaming pipeline.

\section{Hallucination Network}
\label{sec:method}

\subsection{Network architecture and loss function}
\begin{figure}
    \centering
    \includegraphics[width=0.49\textwidth]{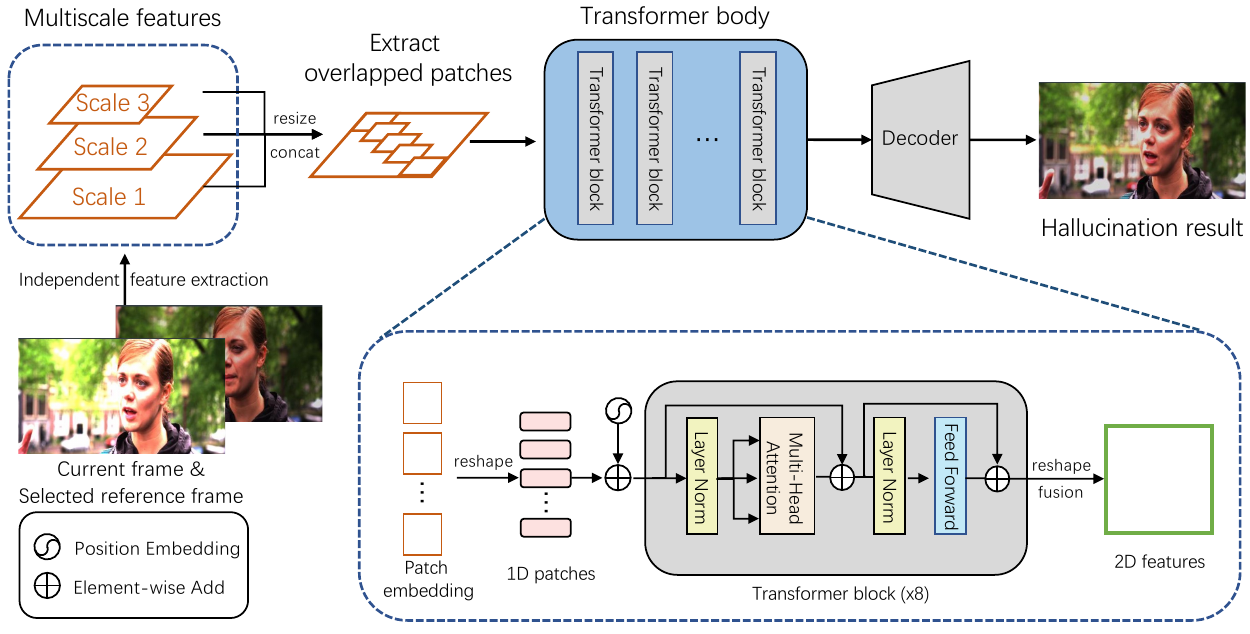}
    \caption{The proposed DNN receives the current and reference frames. It extracts features from overlapping patches at three different scales, then concatenates and processes those with a  transformer with 8 stages to hallucinate HDR details in the overexposed regions of the input frame.}
    \label{fig:network_arch}
\end{figure}

Our transformer-based DNN (Fig.~\ref{fig:network_arch}) takes as input the concatenation of the current ($i_C^{LDR}$) and reference ($i_R^{LDR}$) frame\footnote{Our DNN architecture natively supports multiple reference frames in input, but we investigate the case with one reference frame only.}; from this, it extracts a sequence of feature embeddings. 
We avoid linear projection layers (commonly adopted since ViT~\cite{ViT}) because of their memory consumption and sub-optimal performance for dense prediction tasks~\cite{yuan2021hrformer, liang2021swinir}; instead, we use convolutions and flattening. 
Inspired by~\cite{Liu21_FuseFormer}, which suggests non-overlapping patches may introduce blurry edges, we adopt overlapped patch splitting and get a sequence of 1D patch embeddings, $\mathbf{z}_0 =\left[ \mathbf{e}^1 ; \mathbf{e}^2; \cdots ; \mathbf{e}^N\right]$, where $N$ is the number of patches, each of size 7 $\times$ 7, with 4 pixels overlap in each direction.
The embeddings $\mathbf{z}_0$ are passed to a stack of 8 basic transformer blocks, each including a multi-head self-attention (MSA), an MLP feed-forward, and a Layernorm (LN) layer. Thus we have
\begin{eqnarray}
\mathbf{z}_{\ell}^{\prime} & = & \operatorname{MSA}\left(\operatorname{LN}\left(\mathbf{z}_{\ell-1}\right)\right)+\mathbf{z}_{\ell-1},\\
\mathbf{z}_{\ell} & = & \operatorname{MLP}\left(\operatorname{LN}\left(\mathbf{z}_{\ell}^{\prime}\right)\right)+\mathbf{z}_{\ell}^{\prime},
\end{eqnarray}
for $\ell=1 \ldots 8$.
Finally, we composite $\mathbf{z}_8$ into 2D feature maps by summing up the overlapped pixel values, which could be effectively implemented with the fold operator in PyTorch. We perform bilinear resampling to the original input size, and use two convolutional layers with Leaky ReLU as refining layers. One additional convolutional layer generates the final unbounded output, needed for HDR details, whose value may exceed 1.

\textbf{Multiscale (MS) embeddings.} 
Our base architecture processes the data at a single scale.
Although this is computationally efficient, the large majority of the hallucination algorithms adopt MS architectures, a feature that improves understanding and modeling natural images~\cite{Wan21}.
To add MS capabilities to our DNN, we downsample $i_C^{LDR}$ and $i_R^{LDR}$ three times (Fig.~\ref{fig:network_arch}).
The downsampling blocks are stacks of three convolutions and Leaky ReLU with stride 2 in the second layer. We concatenate the three-level features after resizing them to the largest scale.
We analyze the effect of the MS architecture in an ablation study in Section~\ref{sec:results}.

\textbf{Relative position bias (RPB).} The relative position of the semantic elements in both $i_C^{LDR}$ and $i_R^{LDR}$ potentially conveys important information for the hallucination task.
Therefore, we test the effect of adding relative position bias~\cite{Raf22} to each attention head. This is obtained as
%
\begin{equation}
    \operatorname{Att}(\mathbf{Q}, \mathbf{K}, \mathbf{V})=\operatorname{Softmax}\left((\mathbf{Q K}^T+\mathbf{B})/\sqrt{d}\right) \mathbf{V},
\end{equation}
where $\mathbf{Q}$, $\mathbf{K}$ and $\mathbf{V}$ are the query, key and value matrices, $d$ is the dimension of query/key, and $\mathbf{B}$ is the additional relative position matrix. 
We study the effects of RPB in the ablation study in Section~\ref{sec:results}.

\textbf{Mask Only Loss (MOL).} 
Our training cost function includes one adversarial term, $\mathcal{L}_{\mathrm{adv}}$, and one reconstruction error term, $\mathcal{L}_{\mathrm{rec}}$
\begin{equation}
    \mathcal{L} = 
    \lambda_{\mathrm{rec}}\mathcal{L}_{\mathrm{rec}} + \lambda_{\mathrm{adv}}\mathcal{L}_{\mathrm{adv}},
\end{equation}
where $\lambda_{\mathrm{rec}}$ and $\lambda_{\mathrm{adv}}$ weight the relative importance of the two terms. 
The losses of the discriminator $D$ and generator are, respectively:
\begin{eqnarray}
    \mathcal{L}_D&=&\mathbb{E}_{\mathbf{Y}}\left[\log D(\mathbf{Y})+\mathbb{E}_{\hat{\mathbf{Y}}}[\log (1-D(\hat{\mathbf{Y}}))]\right],\\
    \mathcal{L}_{\mathrm{adv}}&=&\mathbb{E}_{\hat{\mathbf{Y}}}[\log (D(\hat{\mathbf{Y}}))],
\end{eqnarray}
where $\hat{\mathbf{Y}}$ is the output of the hallucination DNN, $\mathbf{Y}$ is a real HDR image, and for the discriminator, we use the same architecture described in prior works~\cite{Yan20_sttn, Liu21_FuseFormer}.

As for $\mathcal{L}_{rec}$, some authors~\cite{Liu21_FuseFormer, Yan20_sttn} reconstruct the entire image $\hat{i}_C^{HDR}$ and define it as
\begin{eqnarray}
    \mathcal{L}_{mask} & = & ||(i_C^{HDR} - \hat{i}_C^{HDR})\odot \mathbf{M}||_1, \\
    \mathcal{L}_{out} & = & ||(i_C^{HDR} - \hat{i}_C^{HDR})\odot (1-\mathbf{M})||_1, \\
    \mathcal{L}_{rec} & = & \mathcal{L}_{mask} + \mathcal{L}_{out},
    \label{eq:rec}
\end{eqnarray}
where $\bf{M}$ is a mask indicating the corrupted area and $i_C^{HDR}$ is the ground truth.
The final output is $i_C^{LDR} \odot (1-\mathbf{M}) + \hat{i}_C^{HDR} \odot \mathbf{M}$, but reconstructing the full image may lead to better-learned self-attention.
Since in our approach, we want to favor, whenever possible, cross-attention towards the reference frame more than self-attention within the current one, we also study the following reconstruction loss function, defined as Mask Only Loss (MOL):
\begin{eqnarray}
    \mathcal{L}_{rec} & = & \mathcal{L}_{mask}. 
    \label{eq:mol}
\end{eqnarray}
The adoption of the loss in Eq.~\ref{eq:rec} or Eq. \ref{eq:mol} is investigated in the ablation study.

\begin{figure}[t]
\centering
\small
\begin{tabular}{c@{\hspace{0.9mm}}c@{\hspace{0.7mm}}c@{\hspace{0.7mm}}c@{\hspace{0.5mm}}c@{\hspace{0.7mm}}c@{0.7mm}}
\includegraphics[width=0.31\linewidth]{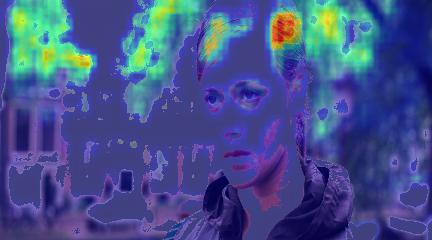}&
\includegraphics[width=0.31\linewidth]{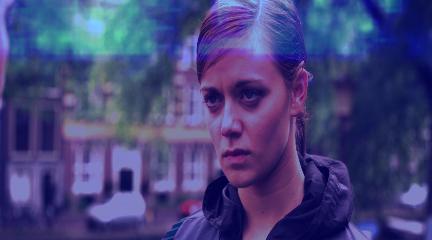}&
\includegraphics[width=0.31\linewidth]{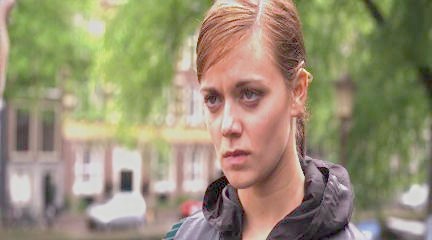}&\\
\includegraphics[width=0.31\linewidth]{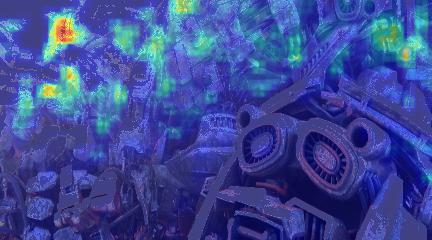}&
\includegraphics[width=0.31\linewidth]{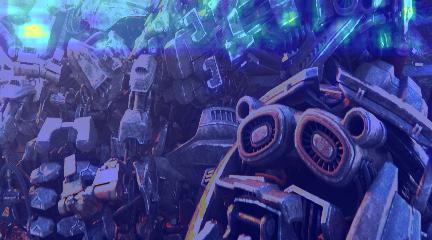}&
\includegraphics[width=0.31\linewidth]{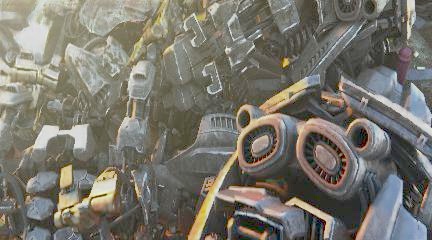}&\\
\includegraphics[width=0.31\linewidth]{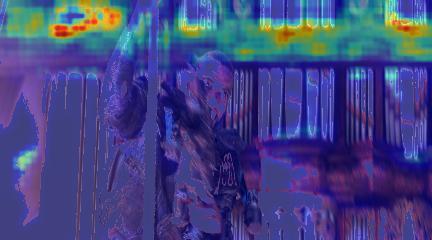}&
\includegraphics[width=0.31\linewidth]{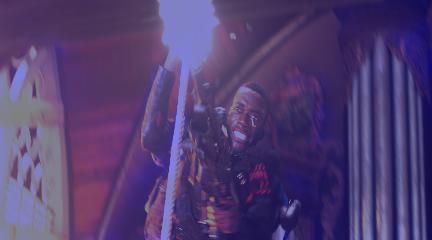}&
\includegraphics[width=0.31\linewidth]{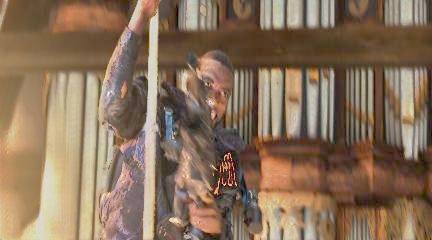}&\\
\small{\textsf{cur.  + att. map}} & \small{\textsf{ref.  + att. map}} & \small{\textsf{output}} \\
\end{tabular}
\vspace{2mm}
\caption{The first two columns show the attention of our hallucination DNN on the current and reference frames. When the current and reference frames share semantic content (first two rows), the attention is spread between the two and generally focuses on complementary areas. The hallucination DNN does not focus on the reference frame when this is too diverse (third row).}
\label{fig:attn_vis}
\end{figure}

\subsection{Hallucination network datasets and training}
\label{sec:hallucination_data}

Large, general datasets for training and evaluation in the HDR field are rare.
Therefore, we train our hallucination DNN in two phases.

We pretrain our network on inpainting on Youtube-VOS~\cite{YoutubeVOS18_b}, a dataset with 5000+ high-resolution LDR videos, 90+ semantic categories, a unique object count exceeding 7800 for more than 340 minutes of length.
We linearize each input RGB image to our hallucination DNN, assuming a $y = x^{1/\gamma}, \gamma = 2$ curve as the camera response function\footnote{We follow~\cite{San20_singlehdr} and use $\gamma = 2$ here.} , \ie{}, we change each channel as $y' = y^{2}$, for $y \in\{R, G, B\}$.
Despite the approximation in the camera response function \cite{Ou22_tonemapping}, our pretrained DNN learns features that are 
relevant for hallucinating HDR details in the linear domain;
we verified experimentally that pretraining leads to higher quality. We use the Adam optimizer for 450K iterations using learning rate 1e-4.

To fine-tune our DNN, we use the Tears of Steel open-source movie~\cite{mango}, containing 145 1920 $\times$ 800 HDR sequences in linear space, which is usually identified as the \emph{Mango} dataset. 
We randomly divide it into 125 training and 20 testing sequences, augment the dataset to sample random 
reference ($i_R^{HDR}$) and current ($i_C^{HDR}$) frame pairs, and
randomly change their exposures\footnote{In the linear HDR domain, a change of exposure is a simple multiplication.} independently from each other.
Since the videos are HDR, we generate the LDR input $i_C^{LDR}$ and $i_R^{LDR}$ by clipping $i_C^{HDR}$ and $i_R^{HDR}$ to the 85th percentile of their combined values.
Finally, we compute the ground truth frame by clipping $i_C^{HDR}$ to the 95th percentile of the combined values of $i_C^{HDR}$ and $i_R^{HDR}$.
This last step is performed as we observed that too large values in the ground truth may cause training to diverge. We fine-tune the network with Adam optimizer for 50K iterations with a learning rate 1e-5.

\section{Adaptive Reference Frame Selection}
\label{sec:frame_selection}

\subsection{Reference frame selection network}

In Table~\ref{tab:nummber_of_ref_framees} we demonstrate that using a single, carefully picked reference frame (instead of a set) can achieve high-quality hallucination at a low compute cost. 
The ideal reference frame selection policy identifies a frame semantically similar to $i_C^{LDR}$, with different exposure, to reveal the details that are poorly visible in $i_C^{LDR}$. For online usage, the selection should use only frames taken from the past.

We formulate the problem as a decision process, solvable by RL~\cite{sutton2018reinforcement}.
We define a system state $s = \{i_{C}^{LDR}, i_{C-1}^{LDR}, i_{C-2}^{LDR}, i_{R}^{LDR} \}$ including the \textbf{history buffer} (the three most recent frames, \ie, the current frame $i_C^{LDR}$ and the previous two, $i_{C-1}^{LDR}$ and $i_{C-2}^{LDR}$) and the \textbf{reference frame buffer} ($i_{R}^{LDR}$ only in our tests, although our method naturally extends to larger reference buffer sizes).
The history buffer in $s$ exposes the recent changes in scene content and exposure to the frame selection DNN, which decides to push $i_C^{LDR}$ into the reference frame buffer\footnote{Other frames from the history buffer could be pushed --- see Section~\ref{sec:results_limitations}.} or not (Fig.~\ref{fig:teaser}), guessing whether $i_C^{LDR}$ is a better candidate than $i_R^{LDR}$ to hallucinate missing details in the next frames. 
We extract features from the frames in $s$ with a pretrained ResNet-18~\cite{He16}. We flatten the features and pass them to a linear layer of size 128, then feed its output to two branches of linear layers.
The first outputs the value function used for training in RL, whereas the second generates the probability of pushing $i_C^{LDR}$ into the reference buffer. The decision to push is made by thresholding at 0.5.

\subsection{Data generation and training}
\label{sec:frame_selection_training}

We train the frame selection DNN using the on-policy, actor-critic A2C~\cite{a2c} in the Stable Baselines 3.0~\cite{stable-baselines3} on a set of 40 parallel environments, with learning rate 7e-5 and other parameters set to their default values.
In each episode, we restore a random sequence of 150 frames from the Mango training set, where we corrupt each sequence to simulate exposure oscillations typical of real systems with autoexposure (see Fig.~\ref{fig:exponential}, details in the Supplementary).
For each frame, we also store the ground-truth HDR frame.
\begin{figure}
    \centering
    \includegraphics[clip=True, trim=0.5cm 7cm 0.5cm 3cm,width=0.49\textwidth]{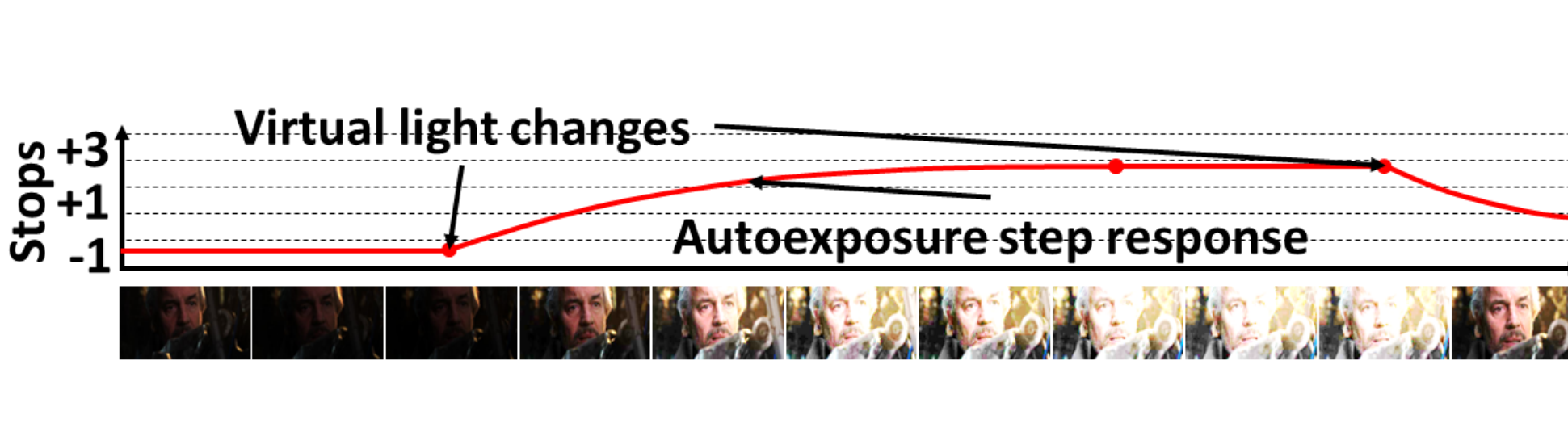}
    \caption{Simulated step response of an autoexposure system. We use this model to create training sequences for the frame selection.}
    \label{fig:exponential}
\end{figure}
In each episode, we initialize the history and reference buffers with the first three frames of the sequence and the third one respectively.
For each frame, we pass then the state $s$ to the frame selection DNN and update (or not) the reference buffer; $i_C^{LDR}$ and $i_R^{LDR}$ are then passed to the hallucination DNN, whose parameters are fixed.
Its output is compared to the ground truth and we use the negative Mean Squared Error (MSE) as a reward.
RL training allows for minimizing the expected MSE.
The policy DNN with the best validation error is finally returned as the output of A2C.

\subsection{Regularize temporal consistency}
\begin{figure}[t]
\centering
\begin{tabular}{c@{\hspace{0.2mm}}c@{\hspace{0.1mm}}c@{\hspace{0.1mm}}c@{\hspace{0.1mm}}c@{\hspace{0.1mm}}}
\rotatebox{90}{\tiny \hspace{1.mm} w/o blending \hspace{-23.5mm} w/ blending}&
\includegraphics[width=0.14\textwidth]{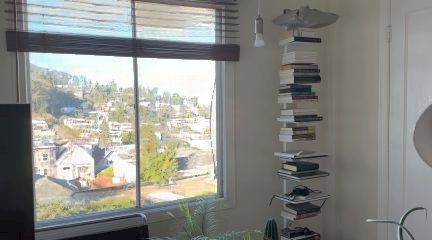}
\includegraphics[width=0.14\textwidth]{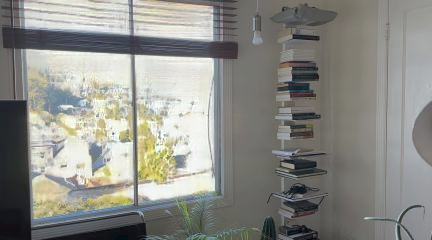}
\includegraphics[width=0.14\textwidth]{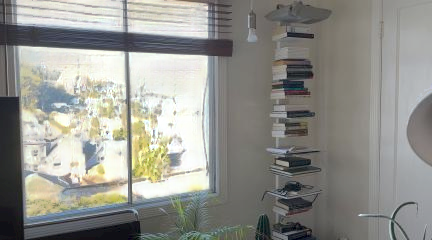}\\
&\includegraphics[width=0.14\textwidth]{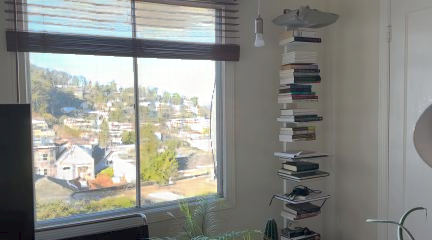}
\includegraphics[width=0.14\textwidth]{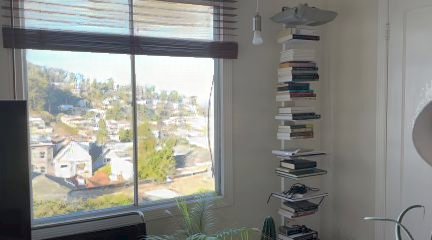}
\includegraphics[width=0.14\textwidth]{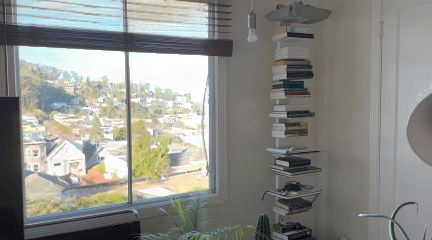}
\end{tabular}
\caption{Three consecutive frames generated without (first row) and with (second row) blending aimed at reducing temporal instability due to a change of reference frame (second column).}
\label{fig:temporal-2}
\end{figure}

HDR video creation requires signal restoration, tone mapping, and temporal stability. 
We regularize the temporal consistency of our method from three aspects: 
(1) Changing the reference frame may cause an abrupt change in the hallucinated area: Fig.~\ref{fig:temporal-2} shows that this is reduced by a simple blending of the outputs generated with the current and past reference frame.
(2) Tone mapping must be consistent for adjacent frames: we are already (and effectively) applying temporal smoothing to the tone mapping parameters.
(3) Many existing temporal consistency methods can also be applied to reduce the jittering artifacts. We provide the details and results in the Supplementary.

\begin{table*}[]
    \centering
    \begin{tabular}{cccccccc}
    \toprule
    Reference frames interval & [t-1, t-1] & [t-1, t-2] & [t-1, t-3] & [t-1, t-4] & [t-1, t-5] & [t-1, t-6] & [t-1, t-7] \\
    \midrule
    PSNR $\uparrow$ & \bf{37.97} & 37.59 & 37.44 & 37.42 & 37.35 & 37.30 & 37.20 \\
    SSIM $\uparrow$ & \bf{0.9812} & 0.9794 & 0.9787 & 0.9787 & 0.9784 & 0.9782 & 0.9778 \\
    \bottomrule
\end{tabular}
\vspace{1mm}
    \caption{Inpainting quality as a function of the reference frame buffer for Fuseformer~\cite{Liu21_FuseFormer}, trained and tested on the Youtube-VOS dataset.}
    \label{tab:nummber_of_ref_framees}
\end{table*}

\begin{figure*}[t]
    \centering
    \hspace{-4mm}
    \small
        \begin{tabular}{
            c@{\hspace{0.7mm}}
            c@{\hspace{0.7mm}}
            c@{\hspace{0.7mm}}
            c@{\hspace{0.7mm}}
            c@{\hspace{0.7mm}}|
            c@{\hspace{0.7mm}}
            c}
             \multicolumn{1}{c}{\small{\textsf{Input image}}} & \multicolumn{1}{c}{\small{\textsf{SingleHDR}}} & \multicolumn{1}{c}{\small{\textsf{Fuseformer}}} & \multicolumn{1}{c}{\small{\textsf{Ours}}} & \multicolumn{1}{c}{\small{\textsf{Ground truth}}} & \multicolumn{1}{c}{\small{\textsf{Reference frame}}} & \multicolumn{1}{c}{\small{\textsf{Mask}}} \\ 
             \includegraphics[width=0.135\linewidth]{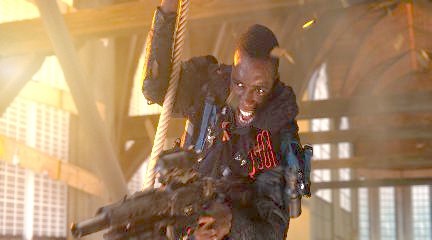}&
            \includegraphics[width=0.135\linewidth]{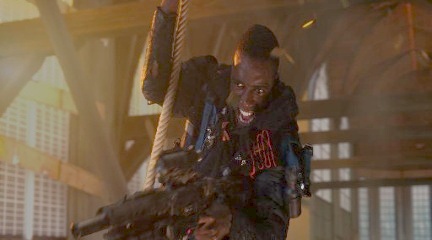}&
            \includegraphics[width=0.135\linewidth]{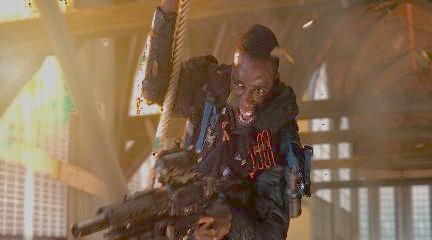}&
            \includegraphics[width=0.135\linewidth]{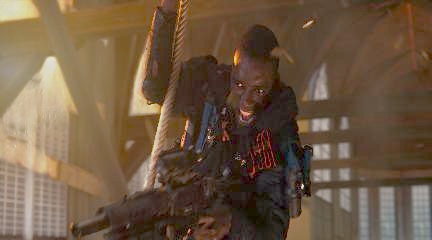}&
            \includegraphics[width=0.135\linewidth]{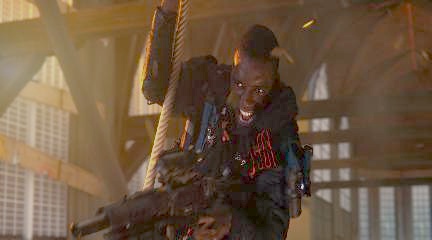}&
            \includegraphics[width=0.135\linewidth]{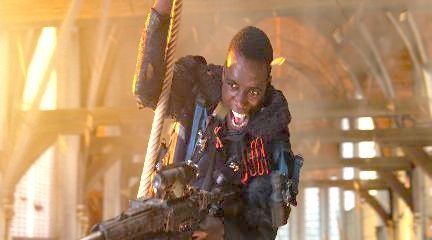}&
            \includegraphics[width=0.135\linewidth]{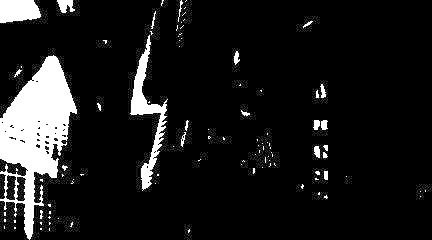}\\ 
             \includegraphics[width=0.135\linewidth]{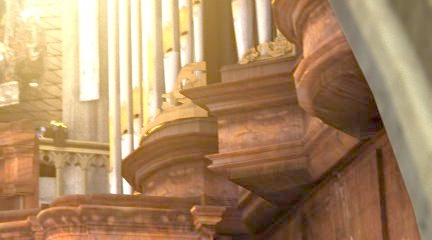}&
            \includegraphics[width=0.135\linewidth]{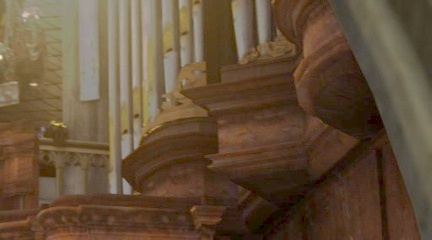}&
            \includegraphics[width=0.135\linewidth]{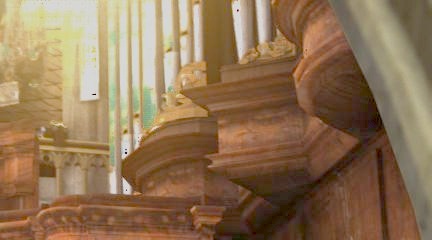}&
            \includegraphics[width=0.135\linewidth]{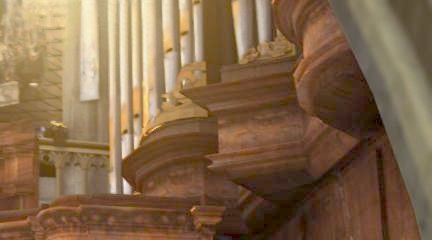}&
            \includegraphics[width=0.135\linewidth]{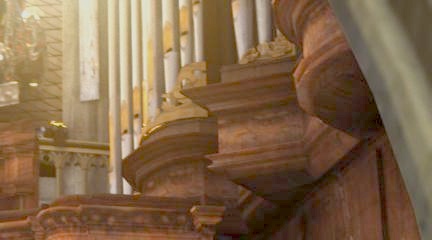}&
            \includegraphics[width=0.135\linewidth]{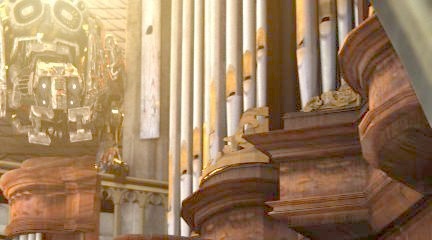}&
            \includegraphics[width=0.135\linewidth]{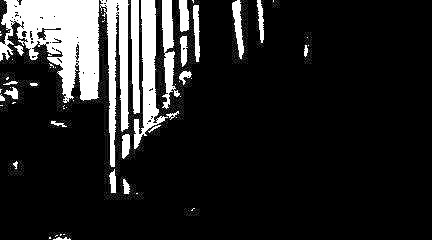}\\ 
             \includegraphics[width=0.135\linewidth]{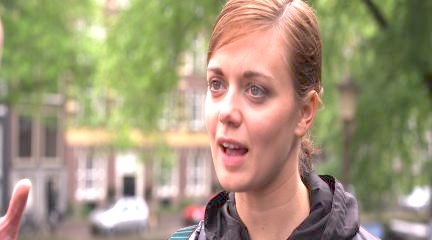}&
            \includegraphics[width=0.135\linewidth]{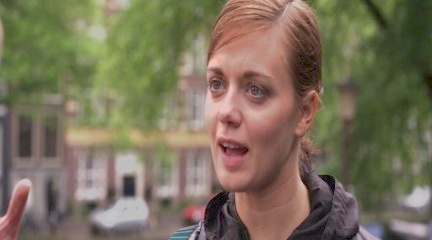}&
            \includegraphics[width=0.135\linewidth]{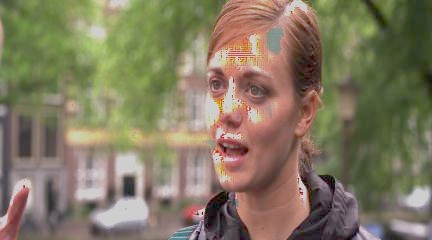}&
            \includegraphics[width=0.135\linewidth]{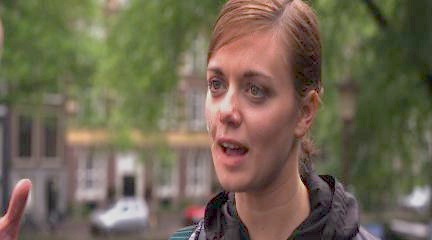}&
            \includegraphics[width=0.135\linewidth]{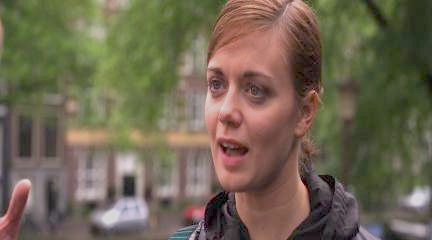} &             \includegraphics[width=0.135\linewidth]{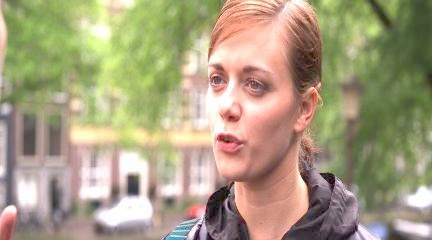}&
            \includegraphics[width=0.135\linewidth]{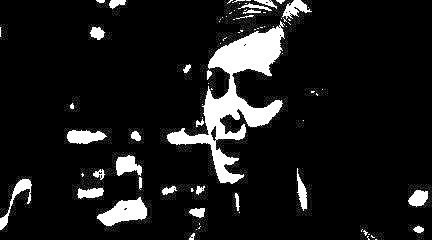}\\
        \end{tabular}
        \caption{Overexposed pixels hallucination results of different algorithms on the Mango test set. The reference frames were selected at $t-10$. The mask shows the overexposed pixels in white.}
    \label{fig:comp_HDR_frames}
\end{figure*} 

\section{Results}
\label{sec:results}
Given the similarities between inpainting and HDR detail hallucination, we test our hallucination DNN on both.
We first demonstrate that one, good reference frame generates comparable inpainting results to a large set of reference frames with a lower computation cost.
We then compare our hallucination DNN against state-of-the-art HDR methods on the overexposed pixels hallucination task. 
We then test the architecture of our model against state-of-the-art inpainting methods.
We show the importance of each component of our architecture on inpainting. 
Finally, we validate our reference frame selection DNN: we quantify the advantage of the learned selection policy against robust baselines on synthetic videos and show the output on synthetic videos.

\subsection{Single vs. multiple reference frames}
We measure the inpainting quality as a function of the number of reference frames on Fuseformer~\cite{Liu21_FuseFormer}, a video inpainting architecture that natively handles a variable number of frames in input.
To this aim, we retrain Fuseformer on the Youtube-VOS training set using causal settings (\ie{}, only past frames are passed in input), and measure the average PSNR and SSIM~\cite{ssim} on the testing set when inputting the reference frame and a set of past frames ranging from the most recent one only to the seven most recent frames.
Table~\ref{tab:nummber_of_ref_framees} shows that using only the most recent frame as reference will not be detrimental for the quality: it can indeed achieve the highest quality in this case. 
This is likely the most informative: because of the temporal vicinity, its content is generally less distorted by movements or occlusions of the subject or the camera, when compared to older frames.
We note that, in any set of reference frames larger than one, this optimal reference frame is always present: despite the attention mechanism, however, Fuseformer does not extract and effectively use the data contained into it.
In summary, isolating and using only the \textit{best} reference frame from a large set seems to be not detrimental for inpainting quality. 
This is exactly the strategy we adopt; it is also the one with the smallest computational cost as it uses only one reference frame.

\subsection{Hallucination network: overexposed pixels hallucination}

\begin{table}[]
\centering
\begin{tabular}{lccc}
\toprule
Method & Ref. frame & \multicolumn{2}{c}{MSE~$\downarrow$} \\
\midrule
SingleHDR  & None & 0.0975 &0.0110 \\
\midrule
Fuseformer & t-1 & 0.1933 & 0.0177\\
Fuseformer & t-10 & 0.1999 & 0.0181\\
\midrule
Ours & t-1 & 0.0519 & 0.0053\\ 
Ours & t-10 & 0.0578 &0.0057 \\ 
\midrule
Overexposure & & severe & mild \\
\bottomrule
\end{tabular}
\vspace{2mm}
\caption{MSE for HDR detail hallucination in the Mango dataset, for severe and mild levels of overexposure.}
\vspace{-4mm}
\label{table:quant}
\end{table}

\begin{figure*}[t]
    \centering
    \hspace{-4mm}
    \small
        \begin{tabular}{
            c@{\hspace{0.5mm}}
            c@{\hspace{0.7mm}}
            c@{\hspace{0.7mm}}
            c@{\hspace{0.7mm}}
            c@{\hspace{0.7mm}}
            c@{\hspace{0.7mm}}|
            c@{\hspace{0.7mm}}
            c}
            \multirow{0}{*}{\rotatebox[origin=c]{90}{Ref t-1 \hspace{0.3cm} Ref t-10\hspace{0.15cm}}}
            &\multicolumn{1}{c}{\small{\textsf{Input image}}} & \multicolumn{1}{c}{\small{\textsf{MAT}}} & \multicolumn{1}{c}{\small{\textsf{FuseFormer}}} & \multicolumn{1}{c}{\small{\textsf{Ours}}} & \multicolumn{1}{c}{\small{\textsf{Ground Truth}}} & \multicolumn{1}{c}{\small{\textsf{Reference Frame}}} & \multicolumn{1}{c}{\small{\textsf{Mask}}}\\
            & \includegraphics[width=0.135\linewidth,clip,trim=0px 0px 1728px 0px]{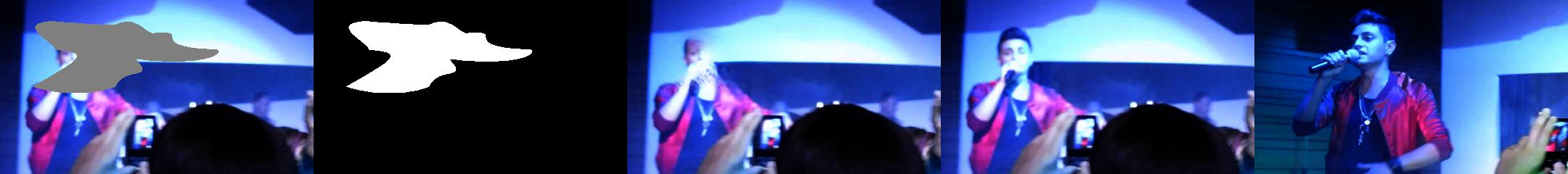}&
            \includegraphics[width=0.135\linewidth]{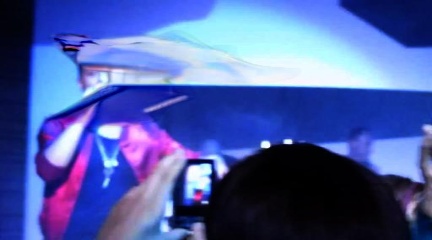}&
            \includegraphics[width=0.135\linewidth]{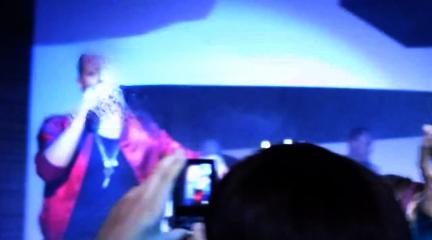}&
            \includegraphics[width=0.135\linewidth]{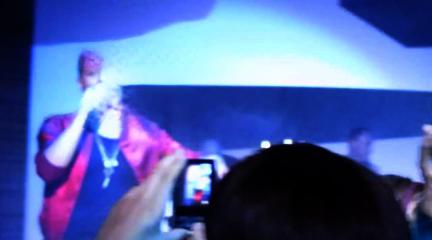}&
            \includegraphics[width=0.135\linewidth]{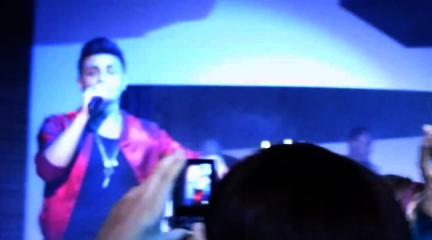}&
            \includegraphics[width=0.135\linewidth,clip,trim=1728px 0px 0px 0px]{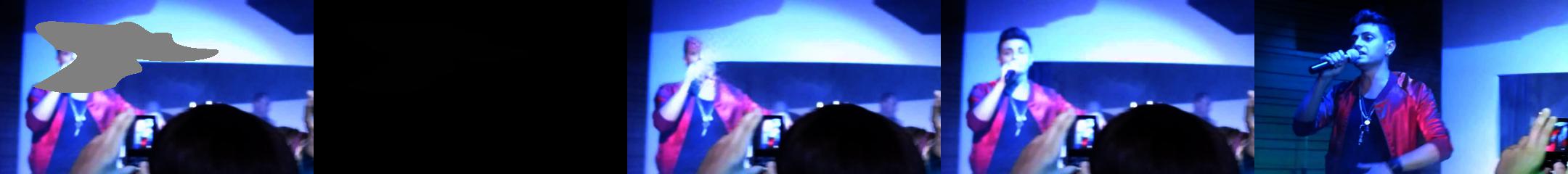} &
            \includegraphics[width=0.135\linewidth]{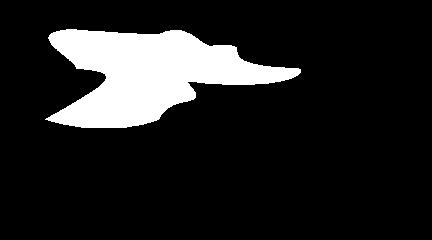}\\
            \multirow{0}{*}{\rotatebox[origin=c]{90}{Ref t-1 \hspace{0.22cm} Ref t-10\hspace{0.14cm}}}
            & \includegraphics[width=0.135\linewidth,clip,trim=0px 0px 1728px 0px]{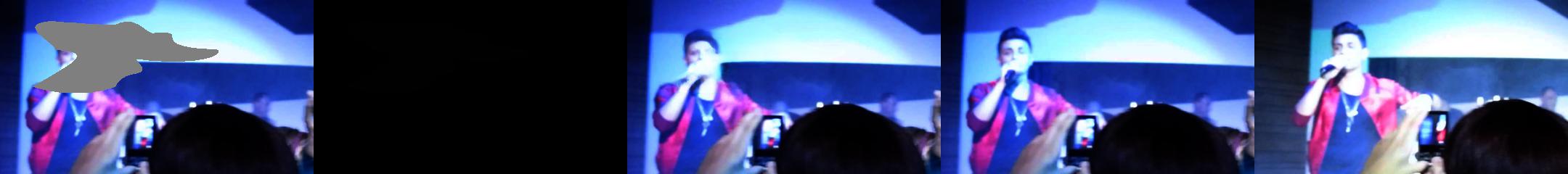}&
            \includegraphics[width=0.135\linewidth]{Figures/regular_inpainting/MAT/gen_1a1dc21969.jpg}&
            \includegraphics[width=0.135\linewidth]{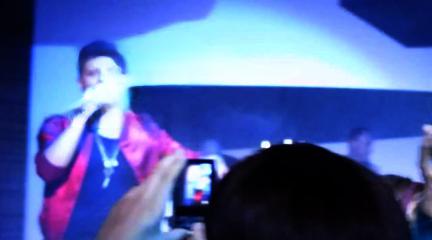}&
            \includegraphics[width=0.135\linewidth]{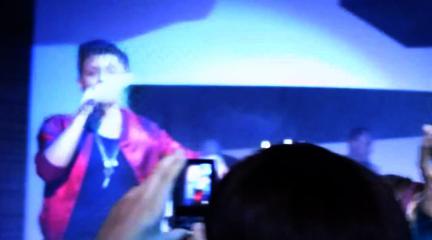}&
            \includegraphics[width=0.135\linewidth]{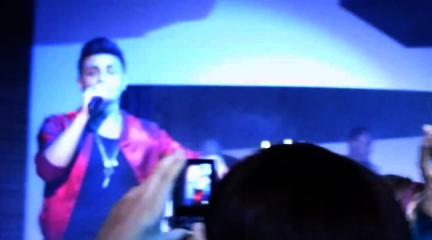}&
            \includegraphics[width=0.135\linewidth,clip,trim=1728px 0px 0px 0px]{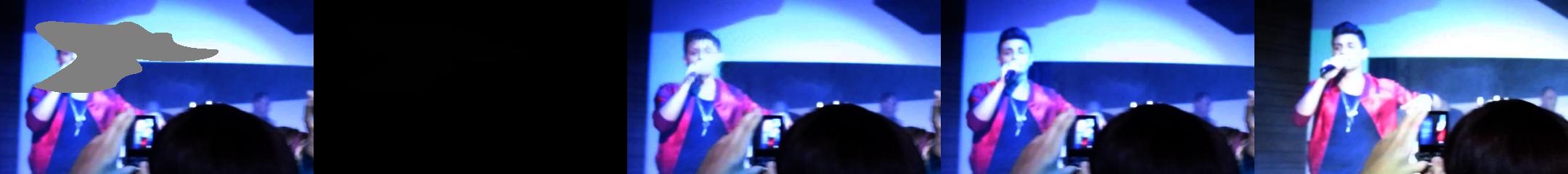} &
            \includegraphics[width=0.135\linewidth]{Figures/regular_inpainting/fuseformer-tminus10/mask_1a1dc21969.jpg}\\
            &\includegraphics[width=0.135\linewidth,clip,trim=0px 0px 1728px 0px]{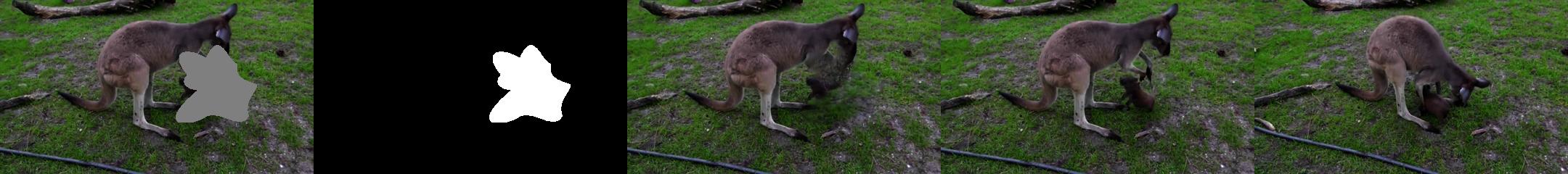}&
            \includegraphics[width=0.135\linewidth]{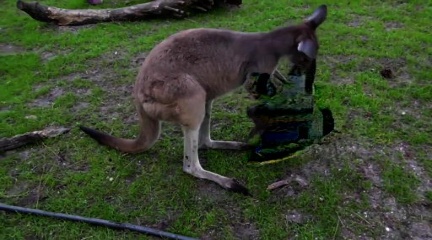}&
            \includegraphics[width=0.135\linewidth]{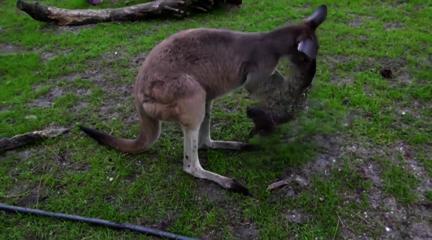}&
            \includegraphics[width=0.135\linewidth]{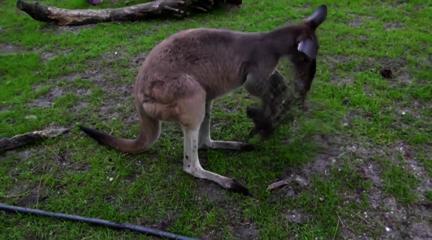}&
            \includegraphics[width=0.135\linewidth]{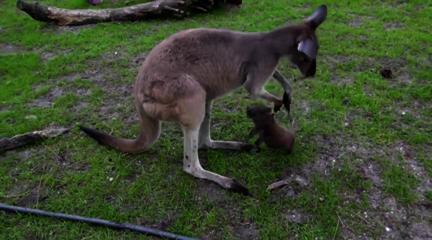}&
            \includegraphics[width=0.135\linewidth,clip,trim=1728px 0px 0px 0px]{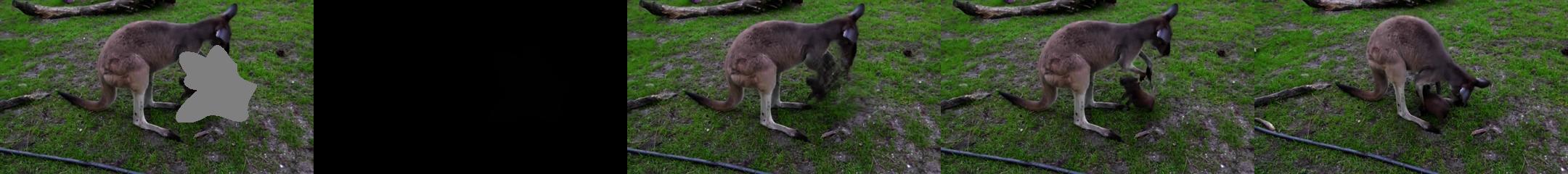} &
            \includegraphics[width=0.135\linewidth]{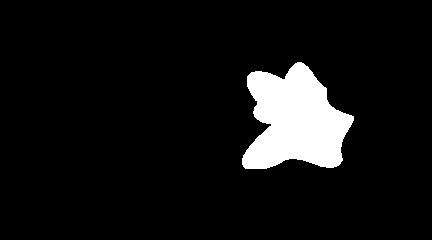}\\
            &\includegraphics[width=0.135\linewidth,clip,trim=0px 0px 1728px 0px]{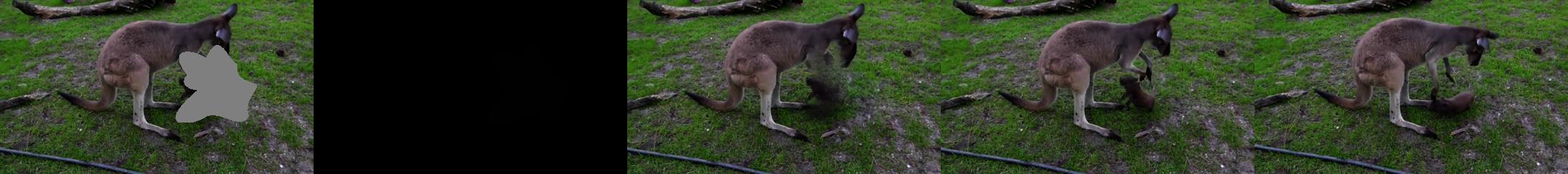}&
            \includegraphics[width=0.135\linewidth]{Figures/regular_inpainting/MAT/gen_1d1601d079.jpg}&
            \includegraphics[width=0.135\linewidth]{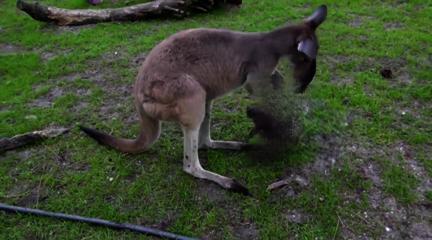}&
            \includegraphics[width=0.135\linewidth]{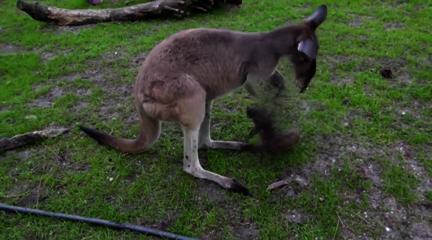}&
            \includegraphics[width=0.135\linewidth]{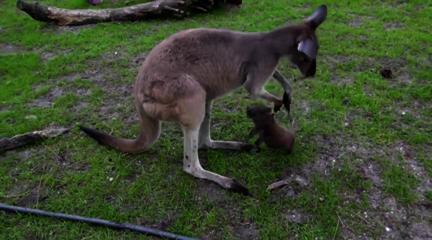}&
            \includegraphics[width=0.135\linewidth,clip,trim=1728px 0px 0px 0px]{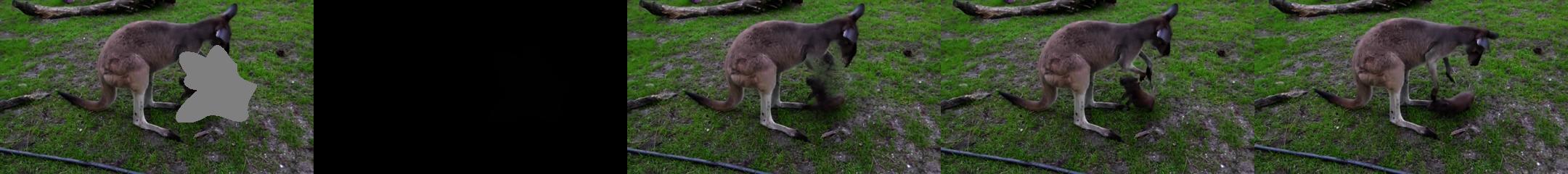} &
            \includegraphics[width=0.135\linewidth]{Figures/regular_inpainting/fuseformer-tminus10/mask_1d1601d079.jpg}\\
        \end{tabular}
        \caption{Inpainting results of different algorithms on the Youtube-VOS test set, performed for a single reference frame taken at distance 1 or 10 from the current one. 
        Our method exceeds MAT~\cite{Li22_mat} visually and it is equivalent or better than Fuseformer~\cite{Liu21_FuseFormer} depending on the scene. These results agree with Table 3 in the main paper. Better reference frames (Ref t-1) leads to better inpainting quality. The mask shows the pixels in the inpainted area in white.}
    \label{fig:regular_inpainting}
\end{figure*}

We compare our hallucination DNN against: Fuseformer using one reference frame, retrained on the Mango dataset; and singleHDR~\cite{San20_singlehdr}, a DNN to hallucinate HDR details using no reference frame, for which we used the DNN parameters shared by the authors.
We pick as reference frames those at distance 1 or 10 from the current one; as for their exposure, we follow the same randomization process described for Youtube-VOS in Section~\ref{sec:hallucination_data}, but we use two different settings where the level of overexpsure of the frames can be severe or mild.
Since the DNNs output HDR images whose values are not in the $[0, 1]$ range, instead of PSNR and SSIM we resort to MSE to quantify the residual error.
Table~\ref{table:quant} reports the average MSE measured on the Mango test set, showing our DNN surpassing the other methods in case of both severe and mild overexposure.
Fig.~\ref{fig:comp_HDR_frames} shows the typical outputs of the algorithms considered here on severe overexposure cases. 
Fuseformer, whose architecture was designed for LDR inpainting, suffers from evident color artifacts. Compared to SingleHDR, our hallucination DNN leverages the reference frame to reconstruct better HDR details in the originally overexposed areas of the input LDR frame.

\subsection{Hallucination network:  inpainting and ablation study}
For single frame inpainting, we compare the architecture of our hallucination DNN against Fuseformer using a single reference frame, both trained on Youtube-VOS. 
The reference frame is at a distance of 1 or 10 frames from the current one, while the exposure of the current and reference frames are randomized as in training (see  Section~\ref{sec:hallucination_data}).
We also compare against MAT~\cite{Li22_mat}, an inpainting DNN that uses no reference frame. We use the MAT parameters shared online, as we found MAT training unstable on Youtube-VOS.
Table~\ref{table:inpainting} reports the average PSNR and SSIM measured on the Youtube-VOS testing set: our architecture overcomes both Fuseformer and MAT in terms of inpainting quality.

\begin{table}[ht!]
\centering
\resizebox{0.47\textwidth}{!}{
\begin{tabular}{lccc}
\toprule
Method & Ref. frame & PSNR (dB)~$\uparrow$ & SSIM~$\uparrow$ \\ \midrule
 MAT~\cite{Li22_mat} & None & 27.49 & 0.940  \\ 
\midrule
 Fuseformer~\cite{Liu21_FuseFormer} & t-1 & 38.86 & 0.983   \\
 Fuseformer~\cite{Liu21_FuseFormer} & t-10 & 35.49 &    0.971   \\
 \midrule
 Ours, MOL & t-1 & 38.88 & 0.984  \\ 
 Ours, MOL + MS & t-1 & \bf{40.30} & \bf{0.987} \\
 Ours, MOL + MS + RPB & t-1 & 40.09 & 0.986 \\ 
 Ours, MOL & t-10 & 35.74 & 0.972  \\ 
 Ours, MOL + MS & t-10 & 36.32 & 0.974 \\
 Ours, MOL + MS + RPB & t-10 & 36.27 & 0.974  \\ 
\bottomrule
\end{tabular}
}
\vspace{2mm}
\caption{PSNR and SSIM for image inpainting on Youtube-VOS, for SingleHDR~\cite{San20_singlehdr}, Fuseformer~\cite{Liu21_FuseFormer} with a single reference frame, and various flavours of the proposed hallucination architecture, including the use of the mask only loss (MOL), the multiscale architecture (MS), and relative position bias (RPB). We use absolute position bias when not using RPB.}
\label{table:inpainting}
\end{table}

Table~\ref{table:inpainting} also shows the PSNR and SSIM for our DNN trained with and without MOL, a multiscale (MS) architecture, and the inclusion of the relative position bias (RPB) encoding. 
Our architecture without MOL, MS, and RPB, is Fuseformer with a single reference frame.
Training using MOL provides +0.02dB over Fuseformer that uses Eq.~\ref{eq:rec} for a reference frame at distance 1, and +0.25dB for distance 10. The adoption of a MS architecture adds +0.58dB for reference frames at distance 1, whereas RPB seems not to provide any advantage.

We provide visual comparisons with Fuseformer and MAT in the Supplementary.

\begin{figure*}[t]
\centering
\includegraphics[width=0.32\linewidth]{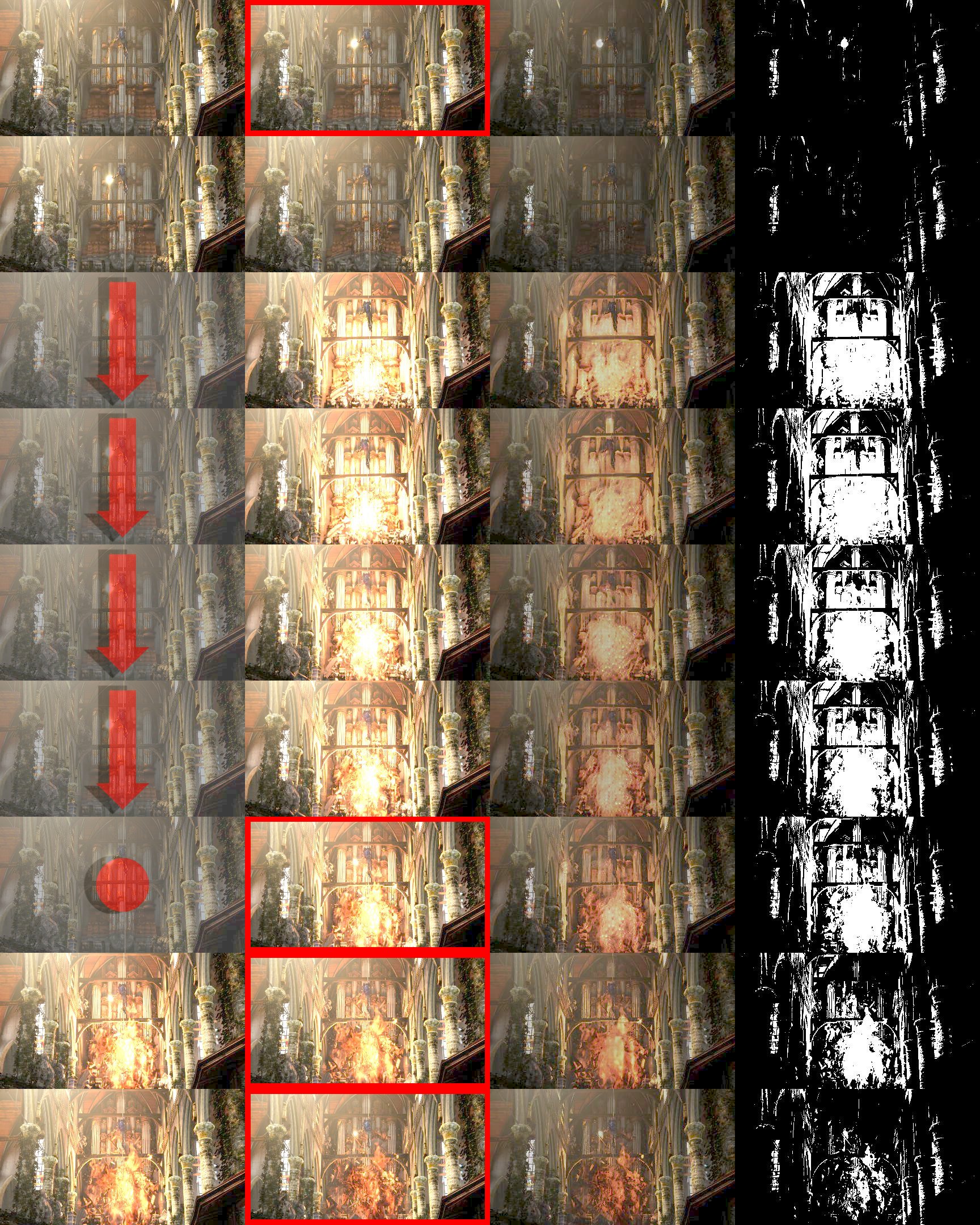}
\includegraphics[width=0.32\linewidth]{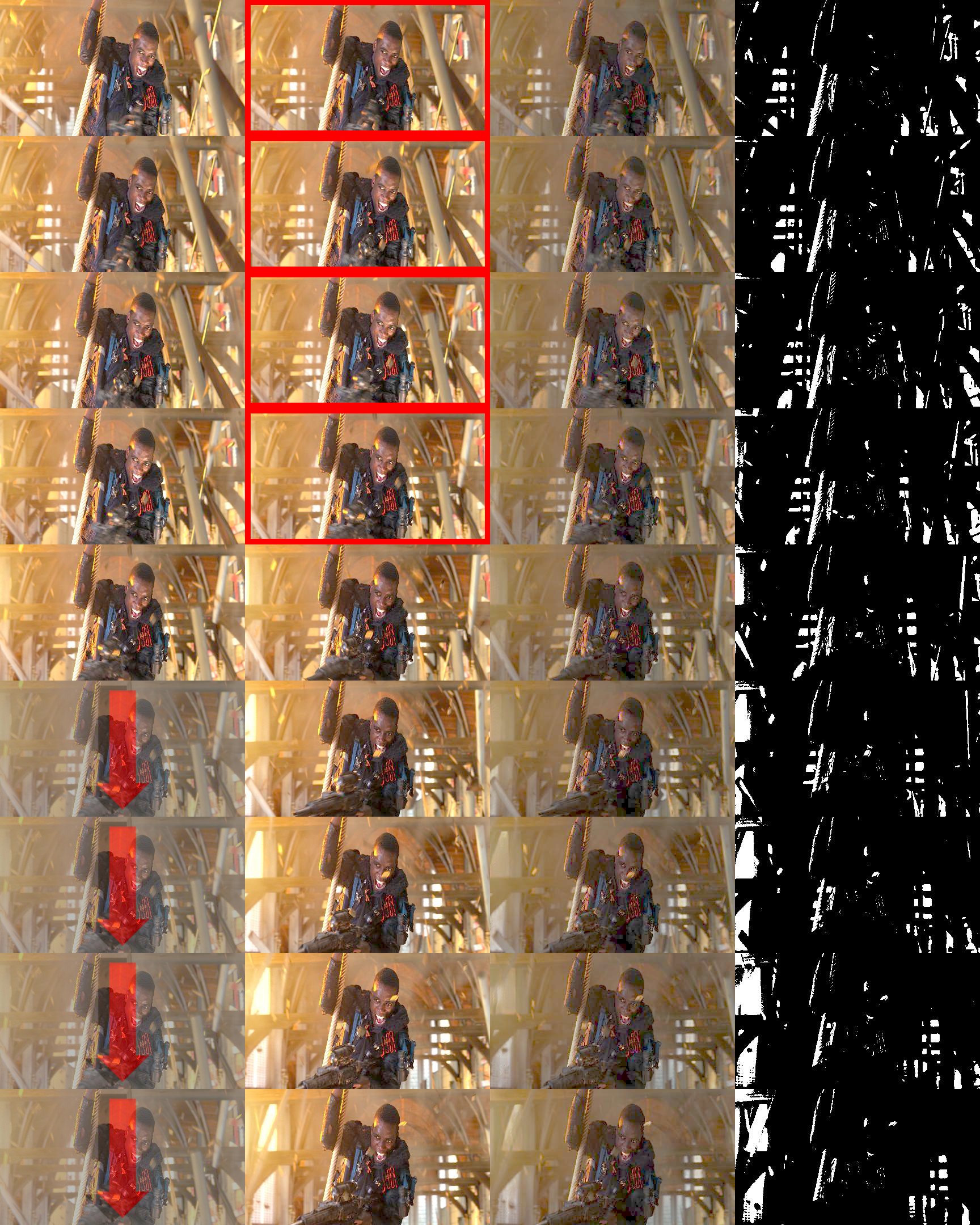}
\includegraphics[width=0.32\linewidth]{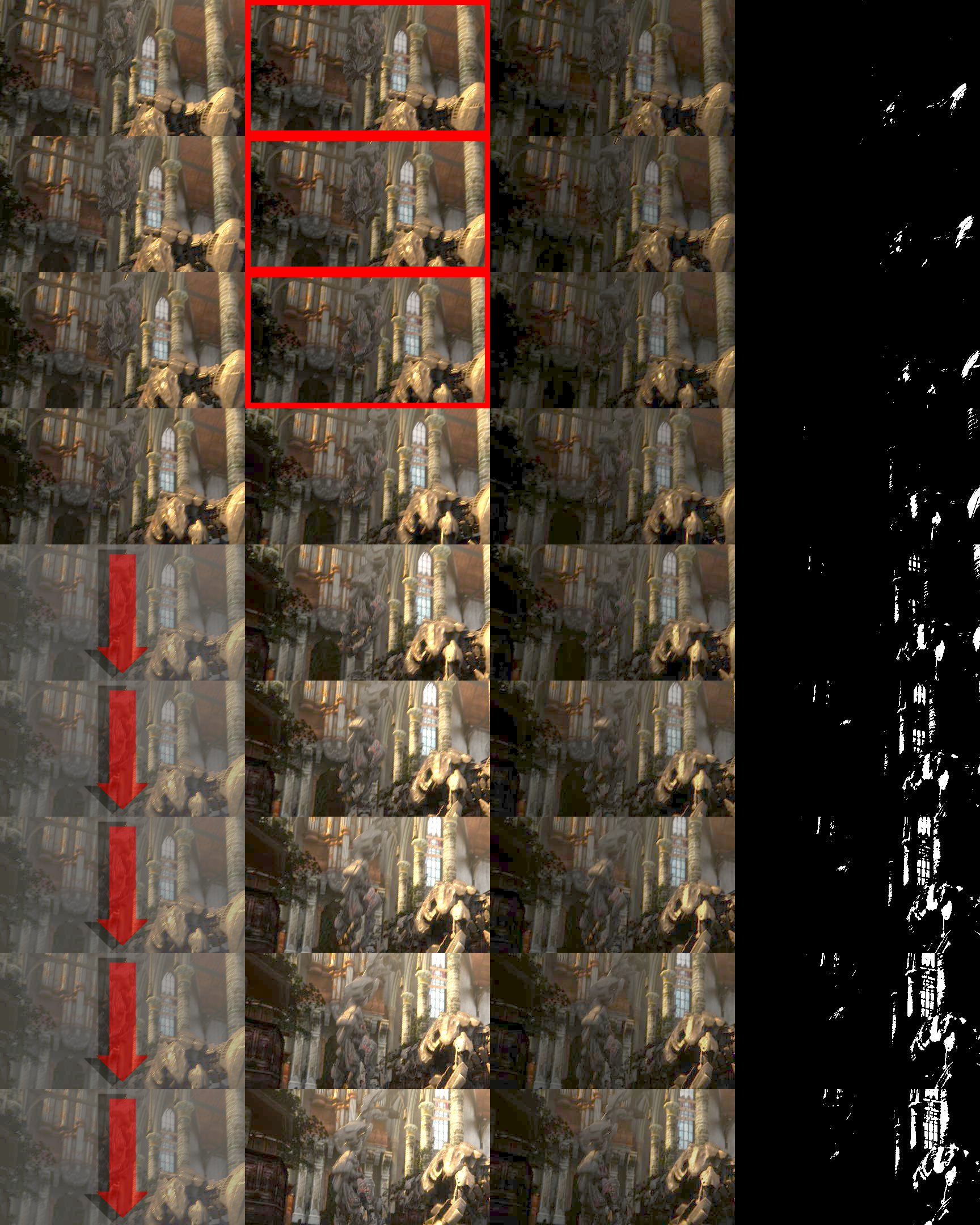} \\
\includegraphics[width=0.32\linewidth]{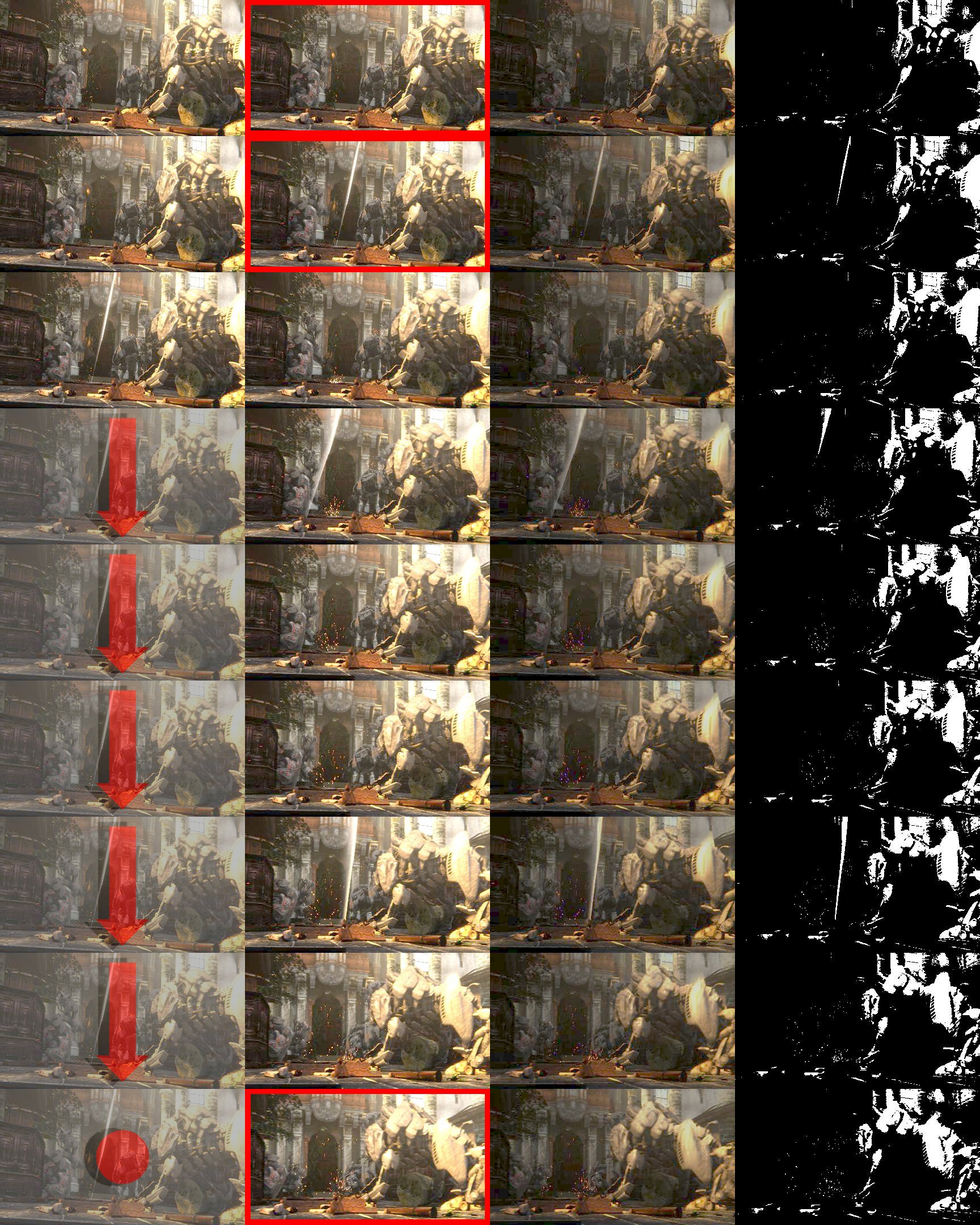}
\includegraphics[width=0.32\linewidth]{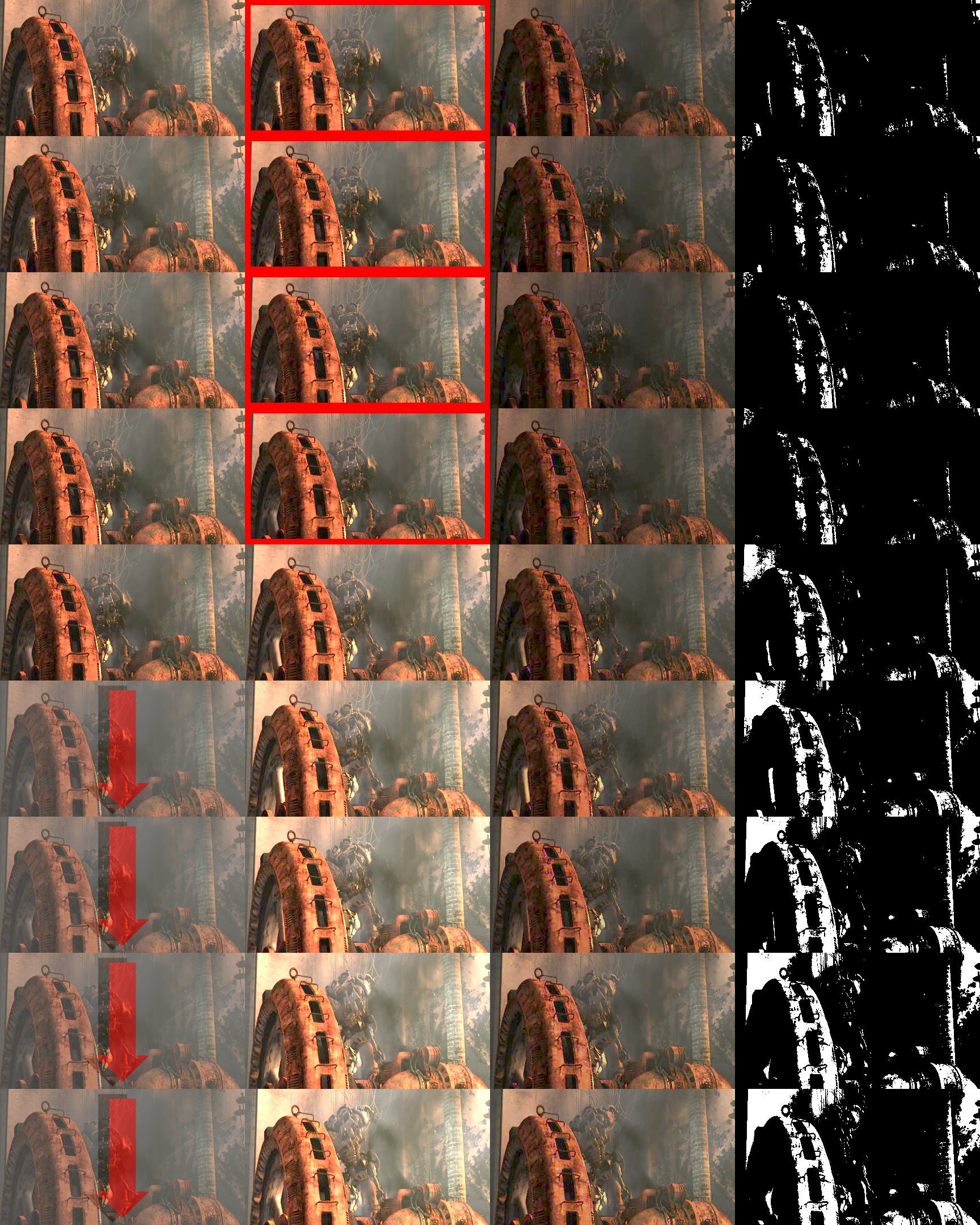}
\includegraphics[width=0.32\linewidth]{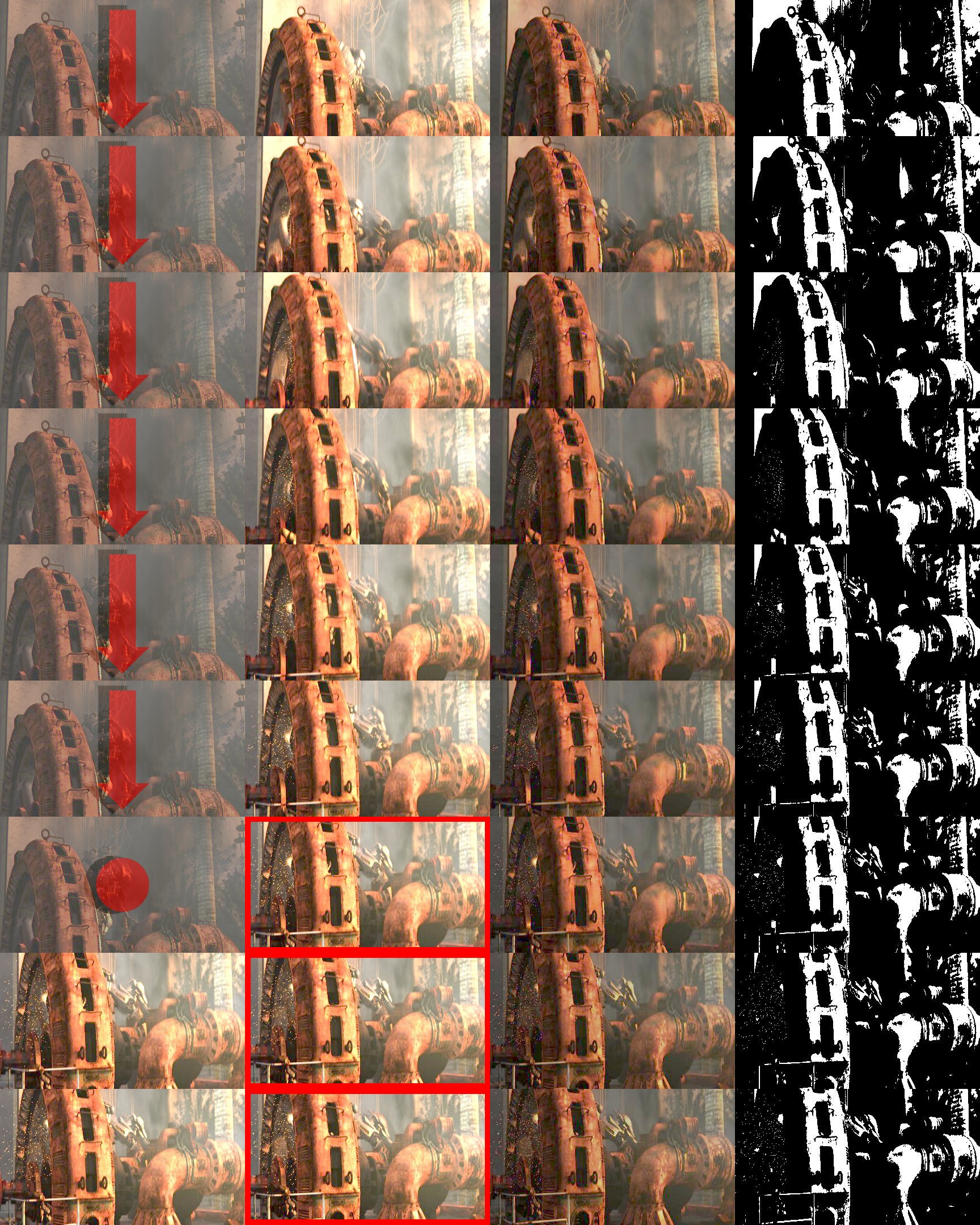}
\caption{Results on synthetic videos from the mango dataset: the first column in each panel shows the reference LDR frame; the arrows indicate that the reference frame in the reference frame buffer is kept unaltered for processing the next frames. The second column is the input LDR frame; the frame is in red if pushed to the reference frame buffer; the third column is the HDR output, tone-mapped with a different gamma for better visualization. The last column shows the saturated pixels in the LDR input. Better seen at $400\%$ zoom.}
\label{fig:comp_synthetic}
\end{figure*}

\subsection{Frame selection}

We evaluate the frame selection DNN on simulated sequences in the Mango testing set, pre-processed as in training (Section~\ref{sec:frame_selection_training}) to simulate real LDR acquisitions.
We compare (in terms of MSE, Table~\ref{tab:frame_selection}) the learned frame selection strategy against a set of baseline policies to push $i_C^{LDR}$ into the reference buffer.
The first baseline (\emph{No reference}) uses a black reference frame, corresponding to pure hallucination; it has the worst quality, as it does not leverage any reference data.
The second baseline (\emph{No push}) is slightly better as it keeps the first frame in the reference frame buffer.
Pushing new frames into the reference frame buffer every 10 frames (\emph{Push every 10 frames} baseline) reduces the MSE, as the reference is now closer in time (and therefore likely semantically more correlated) to the current frame.
Pushing every frame into the reference buffer (\emph{Always push}) further reduces the MSE. But the best baseline (MSE = 0.53) is \emph{Random push}, which can change the reference frame at every step with probability $p=0.5$.
This result is explained considering that the optimal policy should achieve a trade-off between picking a reference frame that is semantically correlated to the current one (and therefore close in time) and one with a widely different exposure (and thus far in time).
This is \emph{on average} achieved by the \emph{Random push} better than other policies, as its average push interval is around 2 or 3 frames.

When compared to these baselines, the learned frame selection policy reduces the MSE to 0.0047, meaning that it performs an adaptive frame selection based on the evidence provided in input that works even better.
Table~\ref{tab:frame_selection} also reports the MSE for an a-posteriori policy based on an explicit search of the best reference frame within the 30 most recent ones: it represents a theoretical upper bound for the MSE that cannot be reached in practice.
The learned policy is significantly closer to this upper bound than the other baselines.
On the other hand, the remaining gap suggests that even better frame selection strategy could exist and be discovered in future.

\begin{table}[]
    \centering
    \begin{tabular}{c@{\hspace{20mm}}c@{\hspace{5mm}}}
    \toprule
    Policy & MSE $\downarrow$ \\
    \midrule
    No reference & 0.0090 \\
    No push & 0.0072 \\
    Push every 10 frames & 0.0065 \\
    Random push (p=0.5) & 0.0053 \\
    Always push & 0.0060 \\
    \midrule
    Learned policy & \bf{0.0047} \\
    \midrule
    A-posteriori (*) & 0.0034 \\
    \bottomrule
    \end{tabular}
    \vspace{2mm}
     \caption{MSE on the Mango testing set for different reference frame selection policies. (*) The a-posteriori policy is a theoretical upper bound that cannot be reached in practice, as it uses the ground-truth frames.}
    \label{tab:frame_selection}
\end{table}

Figure~\ref{fig:comp_synthetic} illustrates the logic of the learned frame selection policy.
When frames darken (underexposure), every frame is pushed into the reference frame buffer; if the exposure later increases, the reference frame is both dark and recent and therefore likely to contains semantically meaningful details that are not overexposed.
For increasing overexposure, no frames are pushed, and the reference frame buffer keeps the last non-overexposed frame.
When the overexposed scene is too different from the reference, however, the current frame is pushed in the reference buffer; the most recent, overexposed frames continue to be pushed, as the result is better using the latest frame as reference, even if overexposed.
Since the policy is learned using RL, the reference frame selection DNN automatically quantifies the semantic closeness and level of overexposure in the current and reference frame and decides to push or not to eventually achieve the optimal trade-off.

\subsection{Computational cost}

On a system equipped with an NVIDIA V100 GPU with 32G of RAM, our hallucination DNN with MS runs @5Hz on $432 \times 240$ images (@14Hz without MS), and inference on the reference frame selection DNN takes 15ms (corresponding to @60Hz).
Although this speed is not yet compatible with real-time execution, our system runs online without requiring to process frames from the future and thus not introducing additional latency; we believe that optimization techniques like pruning~\cite{Mol19},  or Tensor Cores (https://resources.nvidia.com/en-us-tensor-core) may easily accelerate the proposed solution for real time use.

\subsection{Limitations}
\label{sec:results_limitations}

When $i_R^{LDR}$ is widely different than $i_C^{LDR}$ or overexposed in such a way that the same HDR details are lost, the output of the hallucination DNN boils down to pure, no-reference hallucination: it can therefore be plausible, but less adherent to reality.
This may happen in practical situations where the scene is overexposed from the first frame or too rapidly changing.
Such situations, however, can be detected and counteracted as an unbalanced attention between the reference and the current frame, as shown in the last row of Fig.~\ref{fig:attn_vis}.

Another limitation of the approach described here is that, once a reference frame is the reference buffer, the reference frame selection DNN cannot \emph{go back in time} and pick an older frame in the subsequent steps.
This could avoid picking the real best reference frame, leading to a larger MSE; it also makes learning through RL harder.
On the other hand, we can easily allow the reference frame selection DNN to push in the reference frame buffer any frame taken from the history buffer.
Since the DNN is already processing the entire history buffer, this could be done with no or little computational overhead. We plan investigating this in a close future.

Finally, the reference frame buffer could be extended to include more than one frame.
Although this could allow handling difficult situations or keep the memory of past frames for longer, and avoid sudden changes in the output in correspondence of a change of the reference frame, it would also significantly increase the computational cost of the hallucination DNN, that should process a larger set of images and pay the cost (in terms of quality) of a large set of reference frames. It could also incur the problem of lower quality that we measured for large set of reference frames. An alternative could be a hierarchical system devoted to the selection of the reference frame.
Our method also requires a significant speed up to be applied in real-time systems.
\section{Conclusion}
\label{sec:conclusion}

We have introduced a novel architecture for hallucinating missing details in overexposed areas of LDR frames, coupled with a reference frame selection DNN that stores the most promising reference for the future.
Our system puts together problems (and solutions) coming from the fields of inpainting, hallucination, and HDR reconstruction.
It also leverages the temporal oscillations of the exposure of the real LDR acquisition systems to our advantage, to identify frames with different exposures whose details may serve as a reference for hallucinating HDR details in the future.
Our system is designed to work online and therefore can find applications in video-streaming, \eg{} for video conferences, or for mobile video acquisition where the user may frequently and unpredictably change the exposure parameters with a simple click on the screen.
Our analysis shows the benefit of the proposed system but also highlights its limitations, which may be the basis for future research.

{\small
\bibliographystyle{ieee_fullname}
\bibliography{egbib}
}

\end{document}